\documentclass{article}
\usepackage[T1]{fontenc}
\usepackage{iclr2027_conference,times}
\usepackage{amsmath,amssymb}
\usepackage{booktabs}
\usepackage{multirow}
\usepackage{xcolor}
\usepackage{colortbl}
\usepackage{tabularx}
\usepackage{graphicx}
\usepackage{subcaption}
\usepackage{hyperref}
\usepackage{url}
\usepackage{wrapfig}
\usepackage{capt-of}
\usepackage[most]{tcolorbox}
\definecolor{resultrow}{gray}{0.95}

\newcommand{\dashmidrule}{%
  \noalign{\vskip 0.25ex
    \hbox to \linewidth{%
      \hskip\dimexpr 56pt+2\tabcolsep\relax
      \leaders\hbox to 3pt{\hss\vrule height 0.4pt width 1.5pt\hss}\hfill
      \hskip\tabcolsep}%
    \vskip 0.25ex}}
\newcommand{\method}{dSpark}

\iclrfinalcopy
\title{Learning When to Commit: Schedule Distillation for Fast Diffusion Language Models}
\author{Aidar Myrzakhan$^{1}$, Zhiqiang Shen$^{1}$ \\
$^1$VILA Lab, MBZUAI \\
\texttt{\{aidar.myrzakhan,zhiqiang.shen\}@mbzuai.ac.ae} \\
}

\begin{document}
\maketitle

\begin{abstract}
    Diffusion large language models (dLLMs) can generate multiple tokens in parallel, but often require many decoding rounds to complete a response. Efficient generation depends not only on which tokens are easy to predict, but also on how committing to them helps predict the remaining positions. A highly predictable token may provide little useful context, while revealing a less predictable token can help several other positions become predictable sooner. We propose {\bf \em dSpark}, a new schedule distillation-based framework for learning when to commit tokens for faster diffusion language generation. To achieve this, we construct unmasking schedules that prioritize tokens whose reveal makes other positions easier to predict, reducing uncertainty earlier in decoding. We then distill these schedules into the model to enable faster parallel sampling, with a KL-divergence regularizer to help preserve generation quality. We extensively evaluate our approach across mathematical reasoning and code generation tasks. On GSM8K, our method achieves remarkable {\bf 10.12$\times$} and {\bf 5.12$\times$} speedups over the various baselines on LLaDA-8B-Instruct and Dream-7B-Instruct, respectively, while maintaining superior task performance. Our code is available at \url{https://github.com/VILA-Lab/dSpark}.
\end{abstract}

\section{Introduction}

Large language models (LLMs) have achieved remarkable capabilities across natural language understanding~\citep{brown2020language,ouyang2022training,achiam2023gpt,touvron2023llama,grattafiori2024llama,team2023gemini}, and reasoning~\citep{kojima2022large,jaech2024openai,guo2025deepseek}, yet efficient inference remains a fundamental challenge. Most LLMs adopt an autoregressive (AR) paradigm, producing each token conditioned on all previous tokens. This strict left-to-right dependency limits decoding parallelism: generating $L$ tokens typically requires $L$ sequential steps. Diffusion large language models (dLLMs) offer an appealing alternative by generating sequences through iterative denoising. From a partially or fully masked sequence, a dLLM predicts multiple masked positions in parallel and progressively commits them over a series of denoising rounds. Recent models such as LLaDA~\citep{nie2026large} and Dream~\citep{ye2025dream} demonstrate that this paradigm can attain strong language modeling, reasoning, and coding capabilities while enabling flexible generation orders and much greater potential for parallel decoding~\citep{ben2026accelerated,wu2026fast,chen2026dmax}. However, realizing this potential requires reducing the many iterative model evaluations needed to turn a masked sequence into a complete response.

A central factor governing dLLM inference efficiency is the \emph{unmasking schedule}: which token positions should be committed at each decoding round and which should remain masked for subsequent refinement. Given the predictive distributions over all masked positions, a decoder may reveal one token, a fixed number of tokens, or a dynamically selected subset in each iteration. Committing more positions increases parallelism and reduces the number of decoding rounds, but overly aggressive commitments may introduce irreversible errors that contaminate the context used to predict other tokens. Conversely, conservative schedules preserve generation quality but sacrifice much of the computational advantage of parallel prediction. Existing decoding strategies therefore commonly rely on prediction confidence, e.g., committing positions whose maximum token probabilities are highest, under the intuition that highly confident predictions are safer to reveal. While effective as a simple heuristic, this strategy treats each masked position largely by its current local predictability and does not explicitly model how revealing it will change the difficulty of subsequent predictions.

We argue that this distinction between \emph{predictability} and \emph{utility} is crucial for efficient diffusion language generation. The easiest token to predict is not necessarily the most useful token to commit. Consider two masked positions $i$ and $j$. Position $i$ may already have an extremely peaked predictive distribution, yet revealing it may add almost no information about the rest of the sequence. In contrast, position $j$ may be moderately less certain but encode a critical semantic, syntactic, or reasoning decision, once $j$ is revealed, the uncertainty of several other positions may decrease substantially. This is particularly important in structured generation tasks such as mathematical reasoning and code synthesis, where intermediate decisions induce strong dependencies among downstream tokens. A schedule based solely on instantaneous confidence may therefore spend early commitments on safe but informationally redundant tokens, leaving influential positions unresolved for many more rounds. The effective decoding problem is thus not simply \emph{which tokens can be predicted now}, but \emph{which tokens should be committed now so that the remaining sequence becomes easier to predict}.

Motivated by this observation, we propose \emph{dSpark}, a novel framework for learning when to commit tokens during diffusion language generation via {\em schedule distillation}. Our key idea is to construct informative unmasking schedules according to the downstream effect of each commitment rather than its confidence alone. We evaluate candidate commitments by how much revealing them helps reduce predictive uncertainty over the remaining positions. Tokens whose revelation makes multiple unresolved positions easier to determine are prioritized earlier, allowing useful context to propagate more rapidly. Decoding can thus be viewed as a context-construction process: every committed token not only resolves one position but also modifies the prediction problem of all remaining positions. By accounting for this effect, our schedules favor commitments with high \emph{future predictive value}, producing trajectories that reduce sequence-level uncertainty faster than conventional confidence-driven schedules. This enables substantially more useful progress within each expensive model evaluation.

Constructing better schedules alone, however, can add computation if sophisticated token selection must be repeated during inference. We therefore further \emph{distill} the behavior induced by these schedules into the model itself. The constructed schedules provide training trajectories that expose the model to commitment orders in which informative tokens are revealed early and the remaining predictions are progressively simplified. After distillation, the model can exploit these learned dependencies under efficient parallel sampling without expensive search over candidate schedules at every decoding step. Importantly, aggressively optimizing for fast commitment can alter the model's original predictive behavior and degrade generation quality. To mitigate this, we introduce a KL-divergence regularizer that keeps the distilled model close to the original model distribution while learning to operate under accelerated schedules. Schedule distillation thus shifts computation from repeated inference-time schedule optimization into training, yielding a practical mechanism for jointly improving parallelism and preserving the generation capability of the underlying dLLM.

We extensively evaluate {\em dSpark} on two representative models, LLaDA-8B-Instruct and Dream-7B-Instruct, across mathematical reasoning and code generation benchmarks. The results demonstrate that learning \emph{when} to commit is a powerful yet underexplored dimension for accelerating dLLMs. On GSM8K, our approach achieves 10.12$\times$ and 5.12$\times$ speedups over the baselines for LLaDA and Dream, respectively, while maintaining superior task performance. This shows that generation efficiency depends not merely on how many tokens are predicted in parallel, but on how effectively each commitment restructures the prediction problem for later rounds. Our main contributions are:

\begin{itemize}
    \item We identify a key limitation of existing diffusion language model decoding strategies: token confidence alone does not capture the downstream utility of a commitment. We show that revealing a less predictable but more informative token can reduce uncertainty over multiple remaining positions and lead to faster overall decoding.
    \item We propose \emph{schedule distillation}, a new strategy that constructs unmasking schedules by prioritizing tokens according to how much their commitment simplifies future predictions. We further distill these informative schedules into the model, enabling efficient parallel decoding without requiring expensive schedule optimization at inference time.
    \item We extensively validate our method across mathematical reasoning and code generation tasks on multiple dLLMs. On GSM8K, it achieves up to 10.12$\times$ speedup on LLaDA-8B-Instruct and 5.12$\times$ on Dream-7B-Instruct while maintaining strong generation quality.
\end{itemize}
\section{Related Work}

\noindent\textbf{Diffusion Language Models.} Discrete diffusion models generate sequences by reversing a corruption process, and subsequent work developed stronger objectives for language modeling, including SEDD, MDLM, and RADD~\citep{austin2021structured,pmlr-v235-lou24a,sahoo2024simple,ou2025your}. LLaDA \citep{nie2026large} scaled masked diffusion to an instruction-following language model trained from scratch, while Dream~\citep{ye2025dream} demonstrated strong reasoning and coding performance with a model initialized from an autoregressive LLM. Other work combines diffusion within blocks with generation across blocks or scales diffusion models further through conversion from autoregressive models~\citep{arriola2025block,bie2025llada2,zhu2026llada}. These advances establish capable diffusion backbones, but their iterative generation process leaves substantial room to improve inference efficiency.

\noindent\textbf{Efficient Diffusion Language Decoding.} Existing methods reduce either the cost or the number of decoding steps. Fast-dLLM~\citep{wu2026fast}, dKV-Cache~\citep{ma2026dkv}, Elastic-Cache~\citep{nguyen2026attention}, and dInfer~\citep{ma2025dinfer} improve cache reuse and inference execution, while D2F~\citep{wang2026diffusion} and Fast-dLLM v2~\citep{wu2026fast2} use block-based generation to support efficient parallel decoding. Other approaches distill generation into fewer steps or enable more aggressive token updates~\citep{deschenaux2025beyond,pmlr-v267-sahoo25a,chen2026dmax}. Closest to our work, dParallel~\citep{chen2026dparallel} uses certainty-forcing distillation to make tokens ready earlier, whereas d3LLM~\citep{qian2026d3llm} distills pseudo-trajectories and applies multi-block decoding. Training-free approaches such as LEAP~\citep{zhang2026leap} and Ripple-Pivot Search~\citep{ye2026ripple} use lookahead to identify tokens worth committing early. 
\section{Preliminaries}
\label{sec:preliminaries}

\subsection{Masked Diffusion Language Models}
\label{sec:prelim-mdlm}

Masked diffusion language models (MDLMs) generate text by learning to recover tokens from partially masked sequences~\citep{austin2021structured}. For a clean sequence $x_0\in\mathcal{V}^L$, the linear forward process independently replaces each token with the absorbing state
$\mathtt{MASK}$ with probability $t$:
\begin{equation}
\label{eq:forward-masking}
q_{t\mid 0}(x_t\mid x_0)
=\prod_{i=1}^{L}q_{t\mid 0}(x_t^i\mid x_0^i)
=\prod_{i=1}^{L}\operatorname{Cat}\!\left(
  x_t^i;\,(1-t)\delta_{x_0^i}+t\delta_{\mathtt{MASK}}
\right),
\end{equation}
where $\mathcal{V}$ is the token vocabulary, $\mathtt{MASK}\notin\mathcal{V}$, $\operatorname{Cat}$ denotes a categorical distribution, and $\delta_v$ is a point mass at $v$. The noise level $t\in[0,1]$ interpolates between the clean sequence and complete masking. A bidirectional denoiser $p_\theta(x_0^i\mid x_t)$ is trained to reconstruct masked tokens with the reweighted cross-entropy objective~\citep{sahoo2024simple,pmlr-v235-lou24a}
\begin{equation}
\label{eq:mdlm-objective}
\mathcal{L}_{\mathrm{MDM}}(\theta;x_0)
=\int_0^1\frac{1}{t}\,
\mathbb{E}_{q_{t\mid 0}(x_t\mid x_0)}
\left[
  \sum_{i:\,x_t^i=\mathtt{MASK}}
  -\log p_\theta(x_0^i\mid x_t)
\right]\,\mathrm{d}t.
\end{equation}
This formulation underlies large-scale diffusion language models such
as LLaDA~\citep{nie2026large} and Dream-7B~\citep{ye2025dream}, which predict masked tokens in parallel during iterative generation.

\subsection{Parallel Decoding and Unmasking Schedules}
\label{sec:prelim-decoding}

Given a fixed prompt, decoding starts from a fully masked response. Let $s_r$ denote the partially decoded sequence at round $r$ and $\mathcal{M}_r$ the masked response positions eligible for commitment, such as those within an active block or window. A forward pass computes $p_\theta(\cdot\mid s_r,j)$ for each $j\in\mathcal{M}_r$. The decoder selects a nonempty set $S_r\subseteq\mathcal{M}_r$, for example using confidence thresholds~\citep{nie2026large} or predictive-entropy
budgets~\citep{ben2026accelerated}. Under greedy token assignment, the sequence is updated as
\begin{equation}
\label{eq:decoding-update}
s_{r+1}^{j}=\begin{cases}
\displaystyle\operatorname*{arg\,max}_{v\in\mathcal{V}}
  p_\theta(v\mid s_r,j), & j\in S_r,\\[3pt]
s_r^j, & j\notin S_r.
\end{cases}
\end{equation}
An \emph{unmasking schedule} $\pi=(S_0,\ldots,S_{T-1})$ specifies which positions are revealed in each round.

\begin{figure}[!t]
  \centering
  \begin{subfigure}[b]{0.333\linewidth}\centering
    \includegraphics{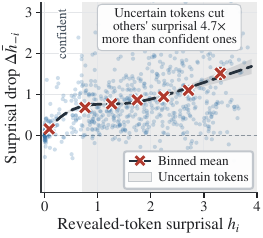}
    \caption{Surprisal drop at other positions}
    \label{fig:motivation-a}
  \end{subfigure}%
  \begin{subfigure}[b]{0.333\linewidth}\centering
    \includegraphics{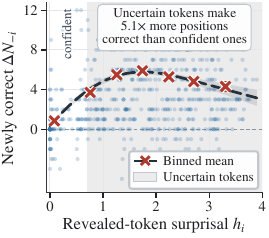}
    \caption{Newly ready positions}
    \label{fig:motivation-b}
  \end{subfigure}%
  \begin{subfigure}[b]{0.333\linewidth}\centering
    \includegraphics{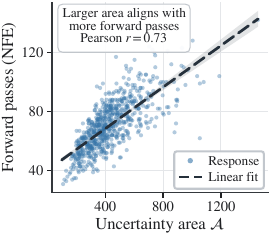}
    \caption{Unc. Area vs.\ forward passes}
    \label{fig:motivation-c}
  \end{subfigure}
  \caption{\textbf{The most predictable token is not the most useful one to reveal.} On correct GSM8K responses from LLaDA, we reveal one masked token $y_i$ and measure its effect on the \emph{other} masked positions. In (a) and (b), each dot is one reveal, and the gray region marks uncertain tokens ($p_{\theta_0}<0.5$), which confidence-based decoders defer.
\textbf{(a)}~Mean surprisal drop at the other positions (nats).
\textbf{(b)}~Net number of other positions that become \emph{ready}, i.e., whose top-1 prediction matches the reference.
\textbf{(c)}~Responses with a larger uncertainty area $\mathcal{A}$ (Eq.~\eqref{eq:uncertainty-area}) need more forward passes.}
  \label{fig:motivation}
\end{figure}

\section{Methodology}
\label{sec:methodology}

\subsection{Motivation}
\label{sec:motivation}

The order in which tokens are revealed can affect how quickly a dLLM completes a response. Each reveal provides context for the remaining positions, but some tokens help more than others. We measure this with the frozen base model $p_{\theta_0}$ on a correct response $y$. In a partially decoded state $s$, the model's uncertainty about the target token at a masked position $j$ is its surprisal $h_j(s)=-\log p_{\theta_0}(y_j\mid s,j)$. Fig.~\ref{fig:motivation-a} and ~\ref{fig:motivation-b} show that revealing less predictable target tokens (larger $h_i$) can reduce uncertainty at other positions and help more predictions match the correct response. We call a position \emph{ready} when the model's most likely token matches the ground truth. Revealing an uncertain token makes $5.1\times$ more other positions ready than revealing a confident one (Fig.~\ref{fig:motivation-b}) and lowers their surprisal $4.7\times$ more (Fig.~\ref{fig:motivation-a}). A useful schedule should therefore consider how each token helps predict the remaining positions, as well as how confident the model is about that token.

\begin{figure}[!t]
  \centering
  \begin{subfigure}[b]{0.45\linewidth}\centering
    \includegraphics{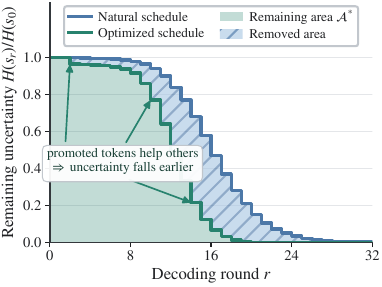}
    \caption{Uncertainty area}
    \label{fig:schedule-construction-a}
  \end{subfigure}%
  \begin{subfigure}[b]{0.45\linewidth}\centering
    \includegraphics{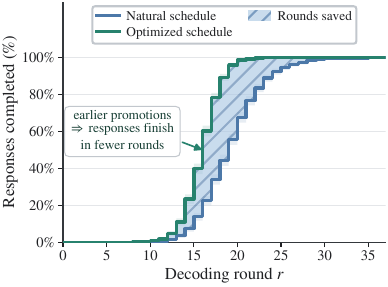}
    \caption{Completion of verified responses}
    \label{fig:schedule-construction-b}
  \end{subfigure}
  \vspace{-0.1in}
  \caption{\textbf{Promoting helpful tokens makes uncertainty fall earlier and responses finish sooner.} Both schedules reveal the same response, in a different order. \textbf{(a)}~The hatched region is the part of the uncertainty area $\mathcal{A}$ (Eq.~\eqref{eq:uncertainty-area}) that the optimized schedule removes. \textbf{(b)}~Share of responses completed by round $r$. The hatched area shows the mean number of rounds saved.}
  \label{fig:schedule-construction}
  \vspace{-0.1in}
\end{figure}

What matters for efficient decoding is not only how much uncertainty a reveal removes, but how early it removes it. Following the masked-token surprisal sum inside the brackets of Eq.~\eqref{eq:mdlm-objective}, we measure remaining reference uncertainty and its accumulated area as
\begin{equation}
\label{eq:uncertainty-area}
H(s_r;y)
  =\sum_{j\in\mathcal{M}_r}h_j(s_r),
\qquad
\mathcal{A}(\pi;y)
  =\sum_{r=0}^{T-1}H(s_r;y).
\end{equation}
The area adds up the model's remaining uncertainty at every decoding round. If revealing a token early makes other tokens easier to predict, their uncertainty is lower in the following rounds and the total area is smaller. Two schedules can therefore produce the same response but have different areas. Figure~\ref{fig:schedule-construction-a} illustrates this: under the natural schedule, the remaining uncertainty stays high for many rounds before it falls, whereas promoting tokens that help the other positions makes it drop earlier and shifts the whole curve to the left, removing the hatched area from $\mathcal{A}$. Since the response is fixed, this gain comes only from the order of reveals. A smaller area also means faster completion. Responses with larger areas under the natural schedule tend to require more forward passes (Fig.~\ref{fig:motivation-c}), and optimized schedules finish verified responses in fewer rounds, never take longer, and gain the most on the responses that need the most rounds (Fig.~\ref{fig:schedule-construction-b}). Our method starts from the natural schedule and searches for tokens whose earlier reveal helps other positions become ready sooner.

\subsection{Uncertainty-Guided Schedule Construction}
\label{sec:schedule-construction}

\begin{figure}[!t]
  \centering
  \includegraphics[width=\linewidth]{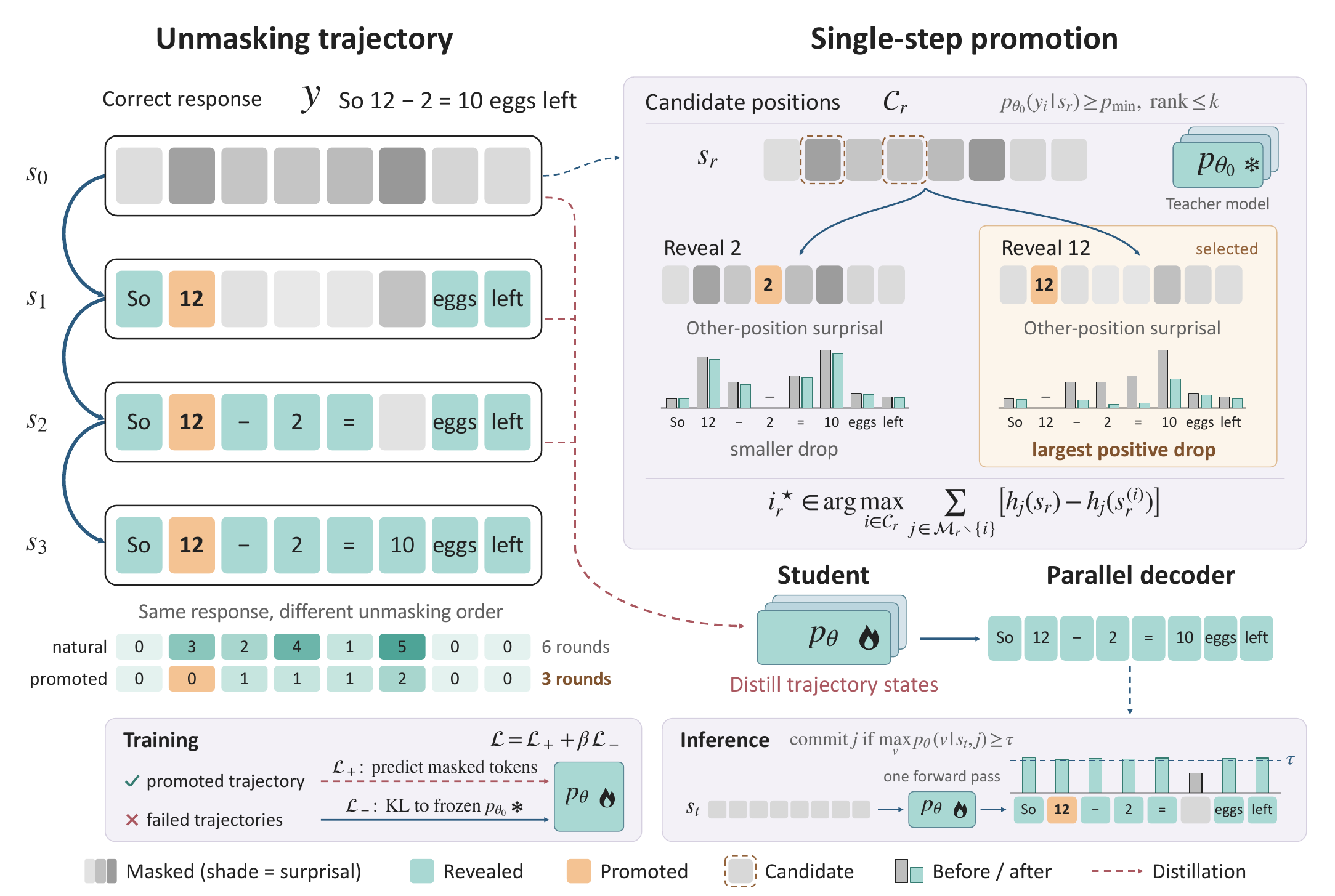}
  \caption{\textbf{Overview of \method{}.} \emph{Left:} for a fixed correct response $y$, the frozen teacher $p_{\theta_0}$ builds an unmasking trajectory that reveals the same tokens in a new order. Masked cells are shaded by surprisal, with darker cells more uncertain. The grid below gives the round in which each token is revealed. Promoting ``12'' first makes the other tokens predictable sooner, so the response is complete after 3 rounds instead of 6. \emph{Right: single-step promotion.} Each plausible candidate $i\in\mathcal{C}_r$ is revealed from the same state $s_r$. Gray and green bars show the surprisal at the other masked positions before and after the reveal. The candidate with the largest positive total drop (Eq.~\eqref{eq:promotion-gain}) is promoted and committed with the ready positions. \emph{Bottom:} the student is trained on promoted trajectories ($\mathcal{L}_+$) with KL to $p_{\theta_0}$ on failed ones ($\mathcal{L}_-$).}
  \label{fig:method-pipeline}
\vspace{-0.1in}
\end{figure}

The uncertainty-area objective in Eq.~\eqref{eq:uncertainty-area} motivates revealing tokens that make the remaining positions easier to predict early in the trajectory. Figure~\ref{fig:method-pipeline} summarizes our framework. Given a correct response $y$, we construct its unmasking trajectory offline with the frozen model $p_{\theta_0}$. The response remains fixed, and we change only the order in which its tokens are revealed. At each round, we compare the candidate reveals by how much they reduce uncertainty at the other masked positions, as illustrated on the right of Figure~\ref{fig:method-pipeline}. This gives a local selection rule based on the context a token provides, rather than only on how predictable that token already is.

At round $r$, we consider plausible target tokens among the masked positions $\mathcal{M}_r$ in state $s_r$,
\begin{equation}
\label{eq:promotion-candidates}
\mathcal{C}_r = \left\{
  i\in\mathcal{M}_r
  \;\middle|\;
  p_{\theta_0}(y_i\mid s_r,i)\ge p_{\min},\quad
  \operatorname{rank}_{\theta_0}(y_i\mid s_r,i)\le k
\right\}.
\end{equation}
Here $p_{\min}$ is a minimum target token probability and $k$ is the maximum allowed rank among vocabulary tokens. For each candidate $i$, let $s_r^{(i)}$ be $s_r$ with $y_i$ revealed and all other positions unchanged. Using the per-position surprisal $h_j(s)=-\log p_{\theta_0}(y_j\mid s,j)$, we select the candidate that most reduces uncertainty at other positions,
\begin{equation}
\label{eq:promotion-gain}
i_r^*\in\operatorname*{arg\,max}_{i\in\mathcal{C}_r}
\sum_{j\in\mathcal{M}_r\setminus\{i\}}
  \left[h_j(s_r)-h_j(s_r^{(i)})\right].
\end{equation}
The score measures the reduction in uncertainty at the other masked positions, excluding the revealed position $i$ itself. We promote the candidate with the largest positive score. If no candidate has a positive score, we skip promotion. Repeating these steps until the response is fully revealed produces $\tau^*(y)=(s_0^*,\ldots,s_T^*)$, where $s_r^*$ is the state after $r$ rounds. This optimized trajectory is illustrated on the left of Figure~\ref{fig:method-pipeline}. We pair its intermediate states $s_r^*$ with the correct response $y$ to train the student $p_\theta$ to predict the remaining masked tokens.

\subsection{Schedule Distillation}

We train the student on intermediate states $s_r^*$ from the optimized trajectory $\tau^*(y)$. The correct response provides the token targets, while the schedule determines which tokens are visible in each training example. The student, therefore, learns to predict the remaining tokens from contexts in which informative positions have been revealed earlier.

Let $\mathcal{D}_+$ denote the distribution of correct responses $y$ and states $s_r^*$ sampled from $\tau^*(y)$. We combine cross-entropy supervision with confidence sharpening. At each state $s_r^*$, $\mathcal{M}_r$ contains the eligible masked positions and $\mathcal{R}_r\subseteq\mathcal{M}_r$ those where the student's most likely token matches the target token $y_j$.
The positive-trajectory objective is
\begin{equation}
\label{eq:schedule-distillation}
\mathcal{L}_+(\theta)
=\mathbb{E}_{(s^*_r,y)\sim\mathcal{D}_+}\!\left[
  \ell_{\mathrm{CE}}(s_r^*,y)
  +\ell_{\mathrm{sharp}}(s_r^*,y)
\right],
\end{equation}
where $\ell_{\mathrm{CE}}$ trains the student to reconstruct the masked tokens of the correct response $y$, using cross-entropy averaged over $\mathcal{M}_r$. Training on intermediate states from our optimized trajectory encourages the student to follow its unmasking progression during decoding. The sharpening term $\ell_{\mathrm{sharp}}$ averages the predictive entropy $\mathcal{H}(p_\theta(\cdot\mid s_r^*,j))$ over currently correct positions $j\in\mathcal{R}_r$, where $\mathcal{H}(q)=-\sum_{v\in\mathcal{V}}q(v)\log q(v)$. Minimizing this term strengthens confidence in correct predictions, supporting earlier commitments during decoding.

The sharpening term, however, cannot tell right from wrong on its own. It is applied only where the student is already correct on the constructed trajectories, but the parameters it updates are shared, so the added confidence carries over to states the student reaches on unsuccessful generation paths. On those states, more confidence can only hurt, so the student must stay where the base model was. We therefore retain failed trajectories for selective KL regularization. Response-level failure does not identify which individual tokens are incorrect, so we use the frozen base model's distribution as the target rather than the failed response's tokens. Let $\mathcal{D}_-$ denote the distribution of state--position pairs sampled for anchoring from these trajectories. The negative-trajectory loss is
\begin{equation}
\label{eq:negative-trajectory-loss}
\mathcal{L}_-(\theta)
=\mathbb{E}_{(s^-,j)\sim\mathcal{D}_-}\!\left[
  D_{\mathrm{KL}}\!\left(
    p_{\theta_0}(\cdot\mid s^-,j)
    \,\Vert\,
    p_\theta(\cdot\mid s^-,j)
  \right)
\right],
\end{equation}
where both models receive the same state $s^-$ and the base model remains frozen. This anchoring limits changes where reliable token-level supervision is unavailable, while leaving the constructed correct trajectories free of the KL penalty. The full training objective is
\begin{equation}
\label{eq:distillation-objective}
\mathcal{L}(\theta)
=\mathcal{L}_+(\theta)+\beta\,\mathcal{L}_-(\theta),
\end{equation}
where $\beta$ controls the regularization strength. Together, these objectives encourage earlier confident commitments for faster parallel decoding while maintaining high task accuracy.

\section{Experiments}
\subsection{Experimental Setup}
\noindent{\textbf{Training Data.}} We build a separate distillation corpus for each backbone. For both models, we generate the trajectories from GSM8K~\citep{cobbe2021training}, PRM12K~\citep{lightman2024let}, NuminaMath-CoT~\citep{numina_math_datasets}, and AceCode~\citep{AceCoder} using a maximum length of $256$ tokens, with a block size of $32$. Overall, the training data contains 52k correct and 38k incorrect samples for LLaDA, and 80k correct and 57k incorrect for Dream. During training, we use all correct trajectories and subsample the incorrect ones to a $4{:}1$ ratio, yielding 52k correct and 13k incorrect trajectories for LLaDA, and 80k correct and 20k incorrect for Dream.

\noindent{\textbf{Implementation Details.}} In our experiments we use LLaDA-Instruct~\citep{nie2026large} and Dream-Instruct~\citep{ye2025dream} models. We fine-tune the models using LoRA~\citep{hu2021lora}. We use a multi-block decoding strategy similar to d3LLM~\citep{qian2026d3llm}. To create a set of target tokens in Eq.~\ref{eq:promotion-candidates}, we use $p_{min}=0.02$ and $k=10$. We set $\beta=1$ in the training objective in Eq.~\ref{eq:distillation-objective}. Training configurations and hardware details are provided in the Appendix.

\noindent{\textbf{Evaluation Details.}} Our evaluation covers mathematical reasoning on GSM8K (0-shot)~\citep{cobbe2021training} and MATH (4-shot)~\citep{hendrycksmath2021}, and code generation on HumanEval (0-shot)~\citep{chen2021evaluating} and MBPP (3-shot)~\citep{austin2021program}. We report accuracy for math and pass@$1$ for code, together with tokens per forward (TPF) and accuracy under parallelism (AUP)~\citep{qian2026d3llm}. The results are averaged over three independent evaluation runs. We use \texttt{lm-eval-harness}~\citep{eval-harness}.

\noindent{\textbf{Baselines.}}
We evaluate our method against competitive dLLM baselines. These include the vanilla LLaDA~\citep{nie2026large} and Dream~\citep{ye2025dream} models and the training-free Fast-dLLM~\citep{wu2026fast}. We also compare with methods that fine-tune the model for faster decoding, namely dParallel~\citep{chen2026dparallel}, d3LLM~\citep{qian2026d3llm}, and D2F~\citep{wang2026diffusion}. 

\subsection{Main Results}

\begin{table}[!t]
\centering
\caption{\textbf{Results by benchmark and backbone.}
Accuracy, tokens per forward (TPF), and accuracy under parallelism (AUP) for LLaDA-8B-Instruct and Dream-7B-Instruct.}
\label{tab:main-results}
\small
\setlength{\tabcolsep}{3pt}
\setlength{\aboverulesep}{0.3ex}
\setlength{\belowrulesep}{0.3ex}
\renewcommand{\arraystretch}{1.03}
\resizebox{1.0\linewidth}{!}{ 
\begin{tabularx}{\linewidth}{
>{\raggedright\arraybackslash}p{56pt}
l
*{6}{>{\raggedleft\arraybackslash}X}}
\toprule[0.8pt]
\multirow{2}{*}{\textbf{Benchmark}} &
\multirow{2}{*}{\textbf{Method}} &
\multicolumn{3}{c}{\textbf{LLaDA-8B-Instruct}} &
\multicolumn{3}{c}{\textbf{Dream-7B-Instruct}} \\
\cmidrule(lr){3-5}\cmidrule(lr){6-8}
& & Acc.$\uparrow$ & TPF$\uparrow$ & AUP$\uparrow$
& Acc.$\uparrow$ & TPF$\uparrow$ & AUP$\uparrow$ \\
\midrule
\multirow{6}{*}{\textbf{GSM8K}}
& Base model   & 78.8 & 1.00 & 78.80  & 82.1 & 1.00 & 82.10 \\
\dashmidrule
& Fast-dLLM~\citep{wu2026fast}   & 75.5 & 2.91 & 199.24 & 80.0 & 1.52 & 122.01 \\
& D2F~\citep{wang2026diffusion}         & 74.2 & 2.83 & 190.87 & 80.2 & 1.71 & 132.39 \\
& dParallel~\citep{chen2026dparallel}    & 75.6 & 7.09 & 469.03 & 82.8 & 3.06 & 246.01 \\
& d3LLM~\citep{qian2026d3llm}      & 74.2 & 9.05 & 571.28 & 80.7 & 4.97 & 379.09 \\
\rowcolor{resultrow}
& \textbf{\method{}}
& \textbf{78.9} & \textbf{10.12} & \textbf{761.67}
& \textbf{83.6} & \textbf{5.12} & \textbf{424.28} \\
\midrule
\multirow{6}{*}{\textbf{MATH}}
& Base model   & 31.7 & 1.00 & 31.70  & \textbf{39.5} & 1.00 & 39.50 \\
\dashmidrule
& Fast-dLLM~\citep{wu2026fast}    & 30.8 & 1.91 & 46.24  & 38.4 & 1.71 & 53.87 \\
& D2F~\citep{wang2026diffusion}          & 28.5 & 2.44 & 45.96  & 38.0 & 1.64 & 51.32 \\
& dParallel~\citep{chen2026dparallel}    & 30.3 & 3.16 & 64.57  & 38.1 & 2.93 & 76.76 \\
& d3LLM~\citep{qian2026d3llm}        & 30.3 & 5.75 & 107.59 & 38.6 & 4.07 & 101.47 \\
\rowcolor{resultrow}
& \textbf{\method{}}
& \textbf{32.0} & \textbf{6.03} & \textbf{119.55}
& 38.9 & \textbf{4.31} & \textbf{106.58} \\
\midrule
\multirow{6}{*}{\textbf{MBPP}}
& Base model   & \textbf{41.7} & 1.00 & 41.70 & \textbf{57.0} & 1.00 & 57.00 \\
\dashmidrule
& Fast-dLLM~\citep{wu2026fast}    & 38.2 & 2.20 & 57.33 & 54.5 & 1.17 & 62.75 \\
& D2F~\citep{wang2026diffusion}          & 38.0 & 1.98 & 50.70 & 53.8 & 1.31 & 67.52 \\
& dParallel~\citep{chen2026dparallel}    & 40.2 & 2.27 & 59.46 & 56.2 & 2.35 & 113.76 \\
& d3LLM~\citep{qian2026d3llm}        & 39.7 & 4.56 & 91.88 & 55.2 & 2.89 & 137.54 \\
\rowcolor{resultrow}
& \textbf{\method{}}
& 40.0 & \textbf{5.52} & \textbf{104.94}
& 56.0 & \textbf{3.77} & \textbf{171.43} \\
\midrule
\multirow{6}{*}{\textbf{HumanEval}}
& Base model   & 38.1 & 1.00 & 38.10 & 55.2 & 1.00 & 55.20 \\
\dashmidrule
& Fast-dLLM~\citep{wu2026fast}    & 37.8 & 2.39 & 52.30 & 54.6 & 1.45 & 66.76 \\
& D2F~\citep{wang2026diffusion}          & 38.2 & 2.71 & 60.59 & 54.0 & 1.39 & 66.78 \\
& dParallel~\citep{chen2026dparallel}    & 39.0 & 4.95 & 83.88 & 54.3 & 2.30 & 91.46 \\
& d3LLM~\citep{qian2026d3llm}        & \textbf{39.6} & 5.95 & 96.58
               & 57.1 & 3.19 & 129.15 \\
\rowcolor{resultrow}
& \textbf{\method{}}
& \textbf{39.6} & \textbf{6.73} & \textbf{106.19}
& \textbf{57.3} & \textbf{4.35} & \textbf{167.17} \\
\bottomrule[0.8pt]
\end{tabularx}
}
\end{table}

Table~\ref{tab:main-results} shows that \method{} achieves the highest TPF and AUP across all four benchmarks on both backbones. The results demonstrate a consistent improvement in the accuracy--parallelism trade-off, with the relative contributions of accuracy retention and increased parallelism varying across tasks.

\noindent{\textbf{Results on LLaDA.}} On GSM8K, \method{} achieves $10.12\times$ the decoding parallelism of the base model while maintaining the base model's accuracy. Compared to d3LLM, it further increases parallelism while closing the accuracy gap relative to the base model, resulting in a $33.3\%$ improvement in AUP. On MATH, \method{} similarly preserves base-model performance while achieving the highest TPF and AUP. The code benchmarks show that the benefits extend beyond mathematical reasoning. On HumanEval, \method{} matches d3LLM's pass@1 while decoding more tokens per forward pass. On MBPP, it improves both pass@1 and TPF over d3LLM. Across all four tasks, \method{} makes better use of parallel decoding than the competing methods, with the highest AUP in every case.

\noindent{\textbf{Results on Dream.}} On GSM8K, \method{} achieves $5.12\times$ the base model's TPF while improving accuracy by $1.5\%$. Relative to d3LLM, the larger parallelism gains occur on code generation: TPF increases by $30.4\%$ on MBPP and $36.4\%$ on HumanEval, accompanied by higher pass@1 on both benchmarks. These improvements give \method{} a substantial AUP lead on the two code tasks and show that its advantage extends across backbones. On MATH, the gain over d3LLM is smaller, but \method{} still achieves the highest TPF and AUP. On MBPP, \method{} achieves the highest TPF and AUP of all compared methods, while also improving pass@1 over d3LLM.

\begin{figure}[!t]
\centering
\includegraphics[width=0.93\textwidth]{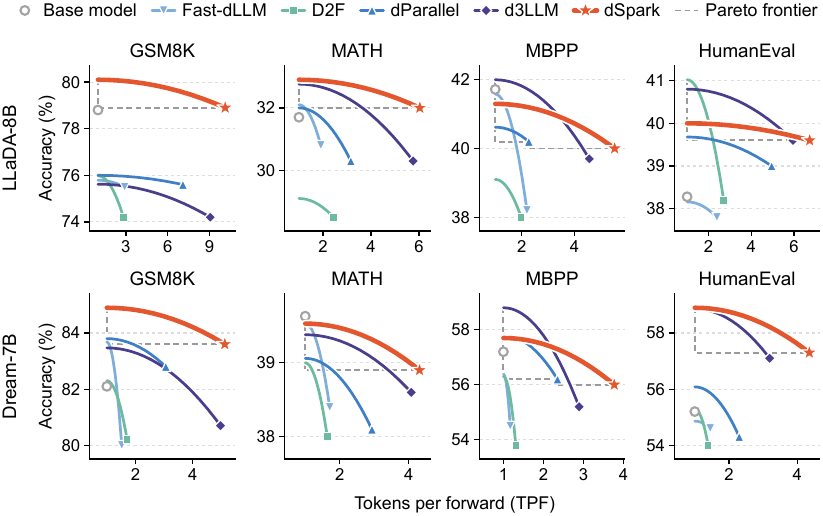}
\vspace{-0.08in}
\caption{\textbf{\method{} extends the accuracy--parallelism frontier on every benchmark and backbone.} Accuracy versus TPF for LLaDA (top) and Dream (bottom). Markers are the operating points of Table~\ref{tab:main-results}. Each curve connects a method's accuracy at TPF $=1$ to its operating point, the two points that make AUP. A flat curve means the method gains parallelism at little cost in accuracy, and a steep drop means it trades accuracy for speed. The dashed gray line is the Pareto frontier.}
\label{fig:pareto}
\vspace{-0.05in}
\end{figure}

\noindent{\textbf{Accuracy--parallelism trade-off.}} Figure~\ref{fig:pareto} shows why \method{} leads in AUP. Methods can decode more tokens per forward pass by accepting a sharp drop in accuracy. \method{} achieves higher TPF while maintaining strong accuracy. This is clearest on GSM8K, where it achieves both the highest TPF and the highest accuracy on LLaDA and Dream. On HumanEval, it exceeds d3LLM's TPF without losing pass@1 on LLaDA and improves both measures on Dream. The other plots show the same overall pattern: \method{} moves decoding toward greater parallelism while preserving enough task performance to make that gain useful. These results are consistent with the goal of \method{}, which encourages the model to reveal informative tokens earlier so that more of the remaining positions can be predicted together. The KL regularizer helps \method{} retain accuracy under aggressive parallel decoding, giving it an advantage over competing methods at higher TPF.

\noindent{\textbf{Confidence and False Positives.}} Distillation for parallel decoding makes a model confident enough to commit many tokens per forward pass. The risk is that it also becomes confident about wrong tokens, which a threshold decoder then commits. Table~\ref{tab:app-calibration} measures this on the natural schedules of 250 test GSM8K samples, recording at every round the prediction at every masked position. The three distilled models are confident about a similar share of correct predictions. Among wrong predictions, however, d3LLM and dParallel are confident for $22.9\%$ and $21.7\%$, while \method{} is confident for $15.5\%$, so its confident predictions are correct $91\%$ of the time, against $82\%$ for the two baselines. Figure~\ref{fig:app-reliability} breaks these predictions into confidence bins. The base model is well calibrated: in every bin, its accuracy is close to its confidence. dParallel and d3LLM are overconfident at every confidence level; for example, their predictions made with a confidence between $0.8$ and $0.9$ are correct only $51\%$ and $52\%$ of the time. \method{} stays much closer to the diagonal, at $65\%$ in the same bin, although it places slightly more predictions in the top bin than the two baselines. Its expected calibration error is also substantially lower than theirs. This more reliable confidence helps explain why \method{} can commit tokens aggressively without the same loss of accuracy.

\subsection{Ablation Study}

\noindent{\textbf{Ablation Study on Training Recipe.}} To examine the contribution of each component, we ablate schedule distillation and KL regularization on LLaDA, with results shown in Table~\ref{tab:ablation}. Using ordinary trajectory training without optimized schedule and KL, it achieves the accuracy of 74.9 and TPF of 7.6 on GSM8K, and 37.9 accuracy and 5.2 TPF on Humaneval. Using the schedule alone substantially increases TPF on both GSM8K and HumanEval, but accuracy declines as the model commits more tokens in parallel. KL regularization alone has the opposite effect: it improves accuracy but reduces TPF, leading to lower AUP than the variant without either component. Combining the two gives the best balance. The full recipe retains 10.1 TPF on GSM8K and 6.7 TPF on HumanEval while raising accuracy to 78.9\% and 39.6\%, respectively. It achieves the highest AUP on both tasks. These results show that the schedule is responsible for the parallelism gain, while KL regularization helps prevent that gain from coming at the expense of accuracy.

\begin{table}[t]
  \centering
  \vspace{-0.1in}
  \small
  \caption{Ablation study on different training recipes of \method{} on the LLaDA-8B-Instruct.}
  \vspace{-0.1in}
  \setlength{\tabcolsep}{6pt}
  \resizebox{0.75\linewidth}{!}{ 
  \begin{tabular}{cccccccc}
    \toprule
    \multirow{2}{*}{\textbf{Schedule}}
      & \multirow{2}{*}{\begin{tabular}[c]{@{}c@{}}\textbf{KL}\\\textbf{Regularization}\end{tabular}}
      & \multicolumn{3}{c}{\textbf{GSM8K}} & \multicolumn{3}{c}{\textbf{HumanEval}} \\
    \cmidrule(lr){3-5} \cmidrule(lr){6-8}
      & & \textbf{TPF} $\boldsymbol{\uparrow}$ & \textbf{Acc} $\boldsymbol{\uparrow}$ & \textbf{AUP} $\boldsymbol{\uparrow}$
      & \textbf{TPF} $\boldsymbol{\uparrow}$ & \textbf{Acc} $\boldsymbol{\uparrow}$ & \textbf{AUP} $\boldsymbol{\uparrow}$ \\
    \midrule
     &  & 7.6 & 74.9 & 463.4 & 5.2 & 37.9 & 82.2 \\
    $\pmb{\checkmark}$ &  & \textbf{11.4} & 74.2 & 718.0 & \textbf{8.3} & 35.1 & 105.8 \\
     & $\pmb{\checkmark}$ & 6.3 & 78.0 & 458.4 & 3.7 & \textbf{40.2} & 74.3 \\
    $\pmb{\checkmark}$ & $\pmb{\checkmark}$ & 10.1 & \textbf{78.9} & \textbf{761.7} & 6.7 & 39.6 & \textbf{106.2} \\
    \bottomrule
  \end{tabular}
  }
  \label{tab:ablation}
  \vspace{-0.15in}
\end{table}

\noindent{\textbf{Ablation study on KL regularization data.}}

\begin{wraptable}{r}{0.36\textwidth}
  \centering
  \small
  \vspace{-0.19in}
  \caption{Accuracy and TPF on LLaDA GSM8K with different ratios of correct to incorrect training trajectories.}
  \label{tab:ratio_ablation}
  \vspace{-0.1in}
  \setlength{\tabcolsep}{6pt}
  \begin{tabular}{ccc}
    \toprule
    \textbf{Data Ratio} & \textbf{Acc (\%)} $\boldsymbol{\uparrow}$ & \textbf{TPF} $\boldsymbol{\uparrow}$ \\
    \midrule
    No inc. data               & 74.19          & 11.4          \\
    16\,:\,1              & 75.89          & 11.15         \\
    8\,:\,1               & 76.80          & 10.92          \\
    \textbf{4\,:\,1}      & \textbf{78.94} & \textbf{10.12} \\
    2\,:\,1    & 78.04          & 9.77          \\
    1\,:\,1    & 79.15          & 7.90          \\
    \bottomrule
  \end{tabular}
  \label{tab:kl_ratio}
  \vspace{-0.2in}
\end{wraptable}
To validate our choice of training mixture, we ablate the ratio of correct to incorrect trajectories on LLaDA and evaluate the resulting models on GSM8K, as shown in Table~\ref{tab:kl_ratio}. Without incorrect trajectories, the model reaches 11.4 TPF but only 74.19\% accuracy. Adding incorrect samples improves accuracy, although a larger share generally reduces parallelism. We choose a $4{:}1$ ratio because it retains 10.12 TPF while raising accuracy to 78.94\%. Increasing the share further to $1{:}1$ brings only a small gain in accuracy, while TPF falls to 7.90. At $2{:}1$, both accuracy and TPF are lower than at $4{:}1$, reaching 78.04\% and 9.77, respectively. We also test using a separate, disjoint set of correct trajectories for KL regularization, keeping the same ratio of $4{:}1$. Compared with using incorrect trajectories at the same ratio, accuracy falls from 78.94\% to 77.10\%, and TPF drops by 30.5\%. Together, these results support using incorrect trajectories for KL regularization at a $4{:}1$ ratio, which provides a strong balance between accuracy and parallel decoding.

\section{Conclusion}

In this work, we studied efficient decoding for diffusion large language models from the perspective of \emph{when to commit} predicted tokens. We showed that token confidence alone is insufficient for effective scheduling, since some less predictable tokens can provide more useful context for resolving the remaining sequence. We introduce {\bf dSpark} enabled by \emph{schedule distillation}, which constructs informative unmasking schedules based on the downstream utility of token commitments and distills these schedules into the model for efficient parallel inference. Experiments on mathematical reasoning and code generation show substantial acceleration, including 10.12$\times$ and $5.12\times$ speedups on GSM8K for LLaDA-8B-Instruct and Dream-7B-Instruct, respectively, while maintaining strong task performance. Our results highlight unmasking schedules as an important learnable component of diffusion language modeling and suggest that better commitment strategies can substantially narrow the gap between the theoretical parallelism and practical efficiency of dLLMs.

\section*{Ethics Statement}

This work focuses on improving the inference efficiency of diffusion large language models and does not involve human subjects or the collection of sensitive personal data. Our method does not introduce new model capabilities beyond those of the underlying pretrained models and therefore inherits their existing limitations and potential risks, including biased, harmful, or inaccurate generations. By reducing the number of decoding steps required during inference, the proposed approach can also help lower computational cost and energy consumption.

\section*{Reproducibility Statement}

We provide detailed descriptions of the proposed method, training objectives, decoding procedures, experimental settings, and evaluation protocols in the main paper and supplementary material. We release the implementation code and relevant configurations to facilitate reproducibility.

\section*{AI Use Statement}

Generative AI tools were used only for language polishing and improving the clarity of the manuscript. All technical content, experimental design, analysis, and conclusions were developed and verified by the authors.

\bibliography{references}
\bibliographystyle{iclr2027_conference}

\clearpage
\appendix
\definecolor{exbase}{HTML}{7F7F7F}
\definecolor{exdp}{HTML}{3F7FC6}
\definecolor{exdt}{HTML}{483E8C}
\definecolor{exours}{HTML}{E4572E}
\definecolor{exbad}{HTML}{C62828}
\definecolor{exgood}{HTML}{1B7F3B}
\newtcolorbox{exbox}[2]{enhanced, colback=#2!6, colframe=#2, boxrule=0.5pt, arc=0pt,
  left=4pt, right=4pt, top=3pt, bottom=3pt, toptitle=1.5pt, bottomtitle=1.5pt,
  coltitle=white, fonttitle=\small, title={#1}}
\newcommand{\exbad}[1]{\textcolor{exbad}{#1}}
\newcommand{\exgood}[1]{\textcolor{exgood}{#1}}
\newcommand{\exfoot}[3]{\par\vspace{3pt}Forward passes: #1\quad Tokens: #2\quad #3}
\makeatletter\setlength{\@fptop}{0pt}\makeatother

\section*{Appendix}
\renewcommand{\thesubsection}{\Alph{subsection}}

\subsection{Additional Results}
\label{app:results}

\subsubsection{Wall-Clock Throughput and KV Caching}
\label{app:throughput}

TPF counts forward passes, but a forward pass with a longer context costs more time. We measure wall-clock throughput on NVIDIA A100-SXM4 GPUs with batch size 1. Table~\ref{tab:app-cache} therefore reports the wall-clock throughput of \method{} on LLaDA, without a KV cache and with the prefix and dual caches of Fast-dLLM~\citep{wu2026fast}. \method{} runs with both caches. With the short 0-shot prompt, the caches barely change throughput (253 and 213 tokens/s, against 249 without a cache), since they also increase the number of forward passes (from 24.8 to 29.4 with the prefix cache). With a 5-shot prompt, the prefix cache raises throughput from 92 to 155 tokens/s ($1.7\times$) at similar accuracy. At 512 tokens, the caches lower accuracy by 1.1--2.7 points without raising throughput. Caching is therefore most useful for long prompts, and \method{} reaches 249 tokens/s without it.

We also implement a static KV cache that runs every decoding step as a captured CUDA graph. It keeps the keys and values of the prompt and of committed tokens and, at each step, recomputes only the active window, the block behind it and a rotating slice of 64 cached positions, so that every cached position is refreshed periodically. With this cache, \method{} generates 313 tokens/s at $79.2\%$ accuracy with the 0-shot prompt, 260 tokens/s with the 5-shot prompt and 300 tokens/s at 512 tokens, $1.3$--$2.8\times$ its throughput without a cache, with accuracy within 1.3 points of it (Table~\ref{tab:app-cache}).

\begin{table}[h]
\centering
\small
\caption{\textbf{Wall-clock throughput of \method{} with KV caching} (LLaDA-8B-Instruct, GSM8K, batch size~1). \emph{Prefix} and \emph{Dual} are the caches of Fast-dLLM~\citep{wu2026fast}, and \emph{Static} is our CUDA-graph cache. TPS: tokens per second.}
\label{tab:app-cache}
\begin{tabular}{llccc}
\toprule
\textbf{Setting} & \textbf{KV cache} & \textbf{Acc.}$\uparrow$ & \textbf{TPF}$\uparrow$ & \textbf{TPS}$\uparrow$ \\
\midrule
\multirow{4}{*}{0-shot, 256 tokens} & None & 78.94 & 10.12 & 249 \\
 & Prefix & 77.86 & 8.52 & 253 \\
 & Dual & 78.17 & 7.94 & 213 \\
 & Static (ours) & 79.23 & 9.51 & \textbf{313} \\
\midrule
\multirow{4}{*}{5-shot, 256 tokens} & None & 79.68 & 9.60 & 92 \\
 & Prefix & 80.29 & 8.07 & 155 \\
 & Dual & 80.97 & 7.45 & 142 \\
 & Static (ours) & 79.08 & 8.53 & \textbf{260} \\
\midrule
\multirow{4}{*}{0-shot, 512 tokens} & None & 84.00 & 11.69 & 168 \\
 & Prefix & 81.35 & 7.94 & 153 \\
 & Dual & 82.94 & 7.28 & 160 \\
 & Static (ours) & 82.71 & 9.86 & \textbf{300} \\
\bottomrule
\end{tabular}
\end{table}

Table~\ref{tab:app-cache-systems} shows that dSpark's gains in decoding parallelism translate into faster inference at batch size 1. On GSM8K, it achieves 249 tokens per second (TPS), a $16.6\times$ improvement over the base LLaDA model and 13.2\% higher throughput than d3LLM. It is also the most accurate method in the table, scoring 78.94\% compared with 78.82\% for LLaDA and 74.21\% for d3LLM. Thus, dSpark's higher TPF corresponds to a wall-clock speedup while preserving the backbone's task performance.

\begin{table}[h]
\centering
\small
\caption{Accuracy and wall-clock throughput against the baselines on GSM8K for LLaDA models with batch size 1.}
\label{tab:app-cache-systems}
\begin{tabular}{lccc}
\toprule
\textbf{Model} & \textbf{Acc.}$\uparrow$ & \textbf{TPF}$\uparrow$ & \textbf{TPS}$\uparrow$ \\
\midrule
LLaDA & 78.82 & 1.00 & 15 \\
Fast-dLLM & 75.51 & 2.91 & 94 \\
D2F & 74.24 & 2.83 & 89 \\
dParallel & 75.62 & 7.09 & 165 \\
d3LLM & 74.21 & 9.05 & 220 \\
\method{} & \textbf{78.94} & \textbf{10.12} & \textbf{249} \\
\bottomrule
\end{tabular}
\end{table}

\subsubsection{Generation Length}
\label{app:length}

\begin{table}[h]
\centering
\small
\caption{\textbf{The gains of \method{} hold across generation lengths.} Accuracy (\%) and TPF on LLaDA for maximum generation length $L$. Bold: best in each column.}
\label{tab:app-length}
\setlength{\tabcolsep}{4pt}
\begin{tabular}{@{}llcccccc@{}}
\toprule
& & \multicolumn{2}{c}{$L=128$}
    & \multicolumn{2}{c}{$L=256$}
    & \multicolumn{2}{c}{$L=512$} \\
\cmidrule(lr){3-4}\cmidrule(lr){5-6}\cmidrule(lr){7-8}
\textbf{Benchmark} & \textbf{Method}
& Acc.$\uparrow$ & TPF$\uparrow$
& Acc.$\uparrow$ & TPF$\uparrow$
& Acc.$\uparrow$ & TPF$\uparrow$ \\
\midrule
\multirow{4}{*}{GSM8K}
 & LLaDA & 72.86 & 1.00 & 78.82 & 1.00 & 82.26 & 1.00 \\
 & dParallel  & 69.53 & 6.28 & 75.62 & 7.09 & 80.89 & 8.66 \\
 & d3LLM      & 68.06 & 8.31 & 74.21 & 9.05
              & 78.92 & 10.73 \\
 & \method{}  & \textbf{73.31} & \textbf{8.85}
              & \textbf{78.94} & \textbf{10.12}
              & \textbf{84.00} & \textbf{11.69} \\
\midrule
\multirow{4}{*}{MATH}
 & LLaDA & \textbf{30.86} & 1.00 & 31.74 & 1.00 & 34.26 & 1.00 \\
 & dParallel  & 27.48 & 2.78 & 30.31 & 3.16 & 30.36 & 3.39 \\
 & d3LLM      & 28.62 & 4.93 & 30.34 & 5.75 & 30.46 & 6.28 \\
 & \method{}  & 30.50 & \textbf{4.96}
              & \textbf{32.03} & \textbf{6.03}
              & \textbf{35.64} & \textbf{6.79} \\
\bottomrule
\end{tabular}
\end{table}

\noindent\textbf{Effect of Generation Length.} Table~\ref{tab:app-length} compares performance at maximum generation lengths of 128, 256, and 512 tokens. dSpark achieves the highest TPF in every setting and the highest accuracy in five of six settings. On GSM8K, it matches or exceeds the base LLaDA model's accuracy at every length while increasing TPF from 8.85 at $L=128$ to 11.69 at $L=512$. On MATH, dSpark is more accurate than both accelerated baselines at all three lengths and reaches the highest accuracy overall at $L=256$ and $L=512$. Its advantage is especially clear at $L=512$, where it achieves 35.64\% accuracy and 6.79 TPF, compared with 30.46\% and 6.28 TPF for d3LLM. These results show that dSpark's gains in parallelism persist as the generation length increases, while accuracy remains strong across both tasks.

\subsubsection{General Knowledge}
\label{app:general}

\noindent\textbf{General Knowledge.} We evaluate dSpark on MMLU~\citep{hendrycks2020measuring} and ARC-C~\citep{clark2018think} to check whether training for faster decoding affects performance beyond math and code. As shown in Table~\ref{tab:app-mmlu}, dSpark matches the base LLaDA model's 63.0\% accuracy on MMLU and scores 47.2\% on ARC-C, which is only 0.4 drop from base model. This suggests that dSpark largely preserves the backbone's general knowledge performance.

\begin{table}[h]
\centering
\small
\caption{\textbf{General knowledge is preserved.} Accuracy (\%) of LLaDA-8B-Instruct before and after \method{} training.}
\label{tab:app-mmlu}
\begin{tabular}{lcccc}
\toprule
 & LLaDA  & dSpark \\
\midrule
MMLU & 63.0 & 63.1 \\
ARC-C & 47.6 &  47.2 \\
\bottomrule
\end{tabular}
\end{table}

\begin{table}[t]
\centering
\small
\caption{\textbf{\method{} is less often confident when it is wrong.} Top-1 predictions at every masked position and decoding round along the natural schedules of 250 GSM8K samples.}
\label{tab:app-calibration}
\begin{tabular}{lcccc}
\toprule
 & Base model & dParallel & d3LLM & \method{} \\
\midrule
Correct predictions with $p\geq0.9$ (\%)$\uparrow$ & 51.6 & 74.4 & 74.3 & 76.1 \\
Wrong predictions with $p\geq0.9$ (\%)$\downarrow$ & 2.5 & 21.7 & 22.9 & 15.5 \\
Predictions with $p\geq0.9$ that are correct (\%)$\uparrow$ & 97.6 & 81.7 & 82.2 & 90.8 \\
Expected calibration error (\%)$\downarrow$ & 4.6 & 19.2 & 20.4 & 8.2 \\
\bottomrule
\end{tabular}
\end{table}

\begin{figure}[t]
\centering
\includegraphics[width=\linewidth]{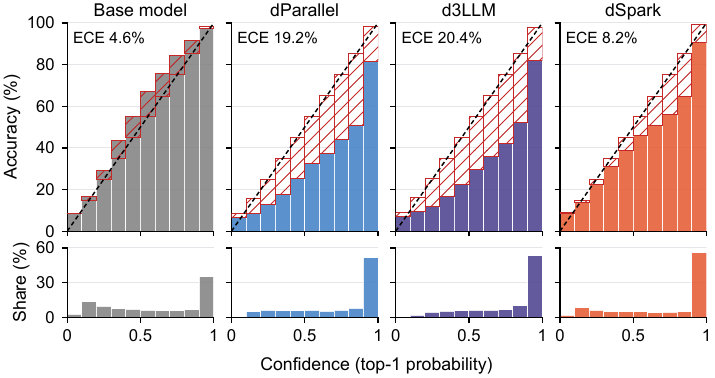}
\caption{\textbf{\method{} stays much closer to calibration than the baselines it is compared with.} Reliability diagrams for the predictions of Table~\ref{tab:app-calibration} in 10 equal-width confidence bins. \emph{Top:} accuracy per bin. The dashed diagonal marks perfect calibration, and the red hatched area is the gap between confidence and accuracy. \emph{Bottom:} share of predictions in each bin.}
\label{fig:app-reliability}
\end{figure}

\subsection{Training Details}
\label{app:training}

We fine-tune both backbones with LoRA~\citep{hu2021lora} on all attention and feed-forward projections, keeping the pretrained weights frozen, and merge the adapters into the backbone before evaluation. Table~\ref{tab:app-training} lists the configuration. The two backbones share every optimization setting and differ only in their training corpora. We train each model on four NVIDIA A100-SXM4 GPUs with data parallelism under DeepSpeed ZeRO stage 2~\citep{9355301}, without CPU offloading.

\begin{table}[h]
\centering
\small
\caption{Training configuration of \method{}.}
\label{tab:app-training}
\setlength{\tabcolsep}{4.5pt}
\begin{tabular}{lcc}
\toprule
 & \textbf{LLaDA-8B-Instruct} & \textbf{Dream-7B-Instruct} \\
\midrule
LoRA rank $r$ / $\alpha$  & \multicolumn{2}{c}{256 / 256} \\
\addlinespace[2pt]
LoRA target modules & \begin{tabular}[c]{@{}c@{}}\footnotesize\texttt{q\_proj, k\_proj, v\_proj,}\\ \footnotesize\texttt{attn\_out, ff\_proj,}\\ \footnotesize\texttt{up\_proj, ff\_out}\end{tabular} & \begin{tabular}[c]{@{}c@{}}\footnotesize\texttt{q\_proj, k\_proj, v\_proj,}\\ \footnotesize\texttt{o\_proj, gate\_proj,}\\ \footnotesize\texttt{up\_proj, down\_proj}\end{tabular} \\
\addlinespace[2pt]
Optimizer & \multicolumn{2}{c}{AdamW~\citep{loshchilov2017decoupled}} \\
Learning rate (schedule) & \multicolumn{2}{c}{$2\times10^{-5}$ (constant)} \\
Weight decay / gradient clipping & \multicolumn{2}{c}{0.01 / 1.0} \\
Epochs & \multicolumn{2}{c}{6} \\
Batch size & \multicolumn{2}{c}{64} \\
Max.\ sequence length & \multicolumn{2}{c}{512} \\
KL weight $\beta$ (Eq.~\ref{eq:distillation-objective}) & \multicolumn{2}{c}{1} \\
Precision & \multicolumn{2}{c}{bfloat16} \\
\bottomrule
\end{tabular}
\end{table}

\subsection{Evaluation Details}
\label{app:evaluation}

\noindent{\textbf{Decoding for \method{}.}} We use a multi-block decoding strategy~\citep{qian2026d3llm} with a confidence threshold of 0.90 to 0.95~\citep{wu2026fast}, which enables decoding both the current and next blocks in parallel. All models decode greedily with at most 256 new tokens and a block size of 32. A new block enters the window once the last active block is $50\%$ decoded and becomes fully active once its predecessor is $50\%$ decoded.  We use no KV cache in the main results.

\noindent{\textbf{Metrics.}} Tokens per forward (TPF) is the number of generated tokens, excluding end-of-sequence padding, divided by the number of forward passes, both summed over the benchmark. Accuracy under parallelism (AUP)~\citep{qian2026d3llm} integrates accuracy over parallelism. For  points $(\rho_1,y_1),\ldots,(\rho_n,y_n)$ sorted by TPF $\rho$, with accuracy $y$, and keeping only points with $y_i\geq y_1-5$,
\begin{equation}
\mathrm{AUP}=\rho_1y_1+\sum_{i=2}^{n}\frac{\rho_i-\rho_{i-1}}{2}\bigl(y_iW(y_i)+y_{i-1}W(y_{i-1})\bigr),
\qquad
W(y)=\min\!\left(e^{-\alpha(1-y/y_{\max})},1\right),
\end{equation}
where $y_{\max}$ is the highest accuracy of any method on the benchmark and $\alpha=3$. Each method contributes its one-token-per-forward point ($\rho=1$) and its reported operating point.

\subsection{Additional Ablations}
\label{app:ablations}

\subsubsection{Candidate Filter}
\label{app:candidate-filter}

A promoted token is committed before the frozen model would commit it on its own, so it is, by construction, a token the model is not yet confident about. The filter in Eq.~\eqref{eq:promotion-candidates} therefore does not ask for confidence. It only asks that the reference token be \emph{plausible} to the model: among its $k=10$ most likely tokens at that position, with a probability of at least $p_{\min}=0.02$, i.e., a surprisal of at most about 4 nats, the range shown in Figure~\ref{fig:motivation}. The two bounds complement each other. When the model's distribution is flat, many tokens exceed $2\%$ and the rank bound applies; when it is peaked, the tenth-ranked token can have a negligible probability and the probability bound applies.

\noindent{\textbf{Why not a stricter filter.}} The tokens that help the other positions most are rarely the confident ones. As shown in Figure~\ref{fig:motivation}, revealing an uncertain token ($p_{\theta_0}<0.5$) makes $5.1\times$ more other positions ready than revealing a confident one and lowers their surprisal $4.7\times$ more. A filter that asks for confidence would therefore exclude the most useful promotions. Promoting these low-probability tokens also does not cost accuracy (Table~\ref{tab:app-branch}).

\noindent{\textbf{Why not a looser filter.}} Below these bounds, the model has effectively ruled the reference token out given the current context. Promoting such a token would make the schedule depend on knowing the verified answer rather than on the model's own uncertainty, and would train the student to commit a token that its context does not support. The bounds also keep the search small, since every candidate is scored with its own forward pass in Eq.~\eqref{eq:promotion-gain}. We set $p_{\min}$ and $k$ from these considerations and did not tune them in end-to-end training.

To test whether low-probability promotions cost accuracy, we take 240 states at which the schedule promotes a token, 80 per probability range, and finish decoding either by committing the promoted token or by committing nothing. Table~\ref{tab:app-branch} shows that committing the promoted token leaves accuracy unchanged within sampling noise while reducing the forward passes needed to finish, by $6.5$ for the least likely tokens.

\begin{table}[h]
\centering
\small
\caption{\textbf{Committing a low-probability promoted token keeps accuracy and saves forward passes.} From 240 states where our schedule promotes a token (80 per probability range), we finish decoding with (\emph{Promoted}) or without (\emph{Not promoted}) committing it.}
\label{tab:app-branch}
\begin{tabular}{lcccc}
\toprule
 & \multicolumn{2}{c}{\textbf{Accuracy (\%)}} & \multicolumn{2}{c}{\textbf{Forward passes to finish}} \\
\cmidrule(lr){2-3}\cmidrule(lr){4-5}
\textbf{Probability of promoted token} & Promoted & Not promoted & Promoted & Not promoted \\
\midrule
$[0.02,0.1)$ & 80.0 & 80.0 & 31.6 & 38.1 \\
$[0.1,0.3)$ & 85.0 & 86.3 & 24.8 & 28.6 \\
$[0.3,1]$ & 86.3 & 83.8 & 25.1 & 27.7 \\
\bottomrule
\end{tabular}
\end{table}

\subsubsection{Token Promotion Strategy.}
We compare the proposed surprisal-reduction criterion Eq.~\eqref{eq:promotion-gain} with the natural schedule, random promotion of a plausible token, and promotion of the highest-surprisal token on GSM8K as shown in Table~\ref{tab:promotion-ablation}. Without KL regularization, our criterion reaches 11.4 TPF, compared with 7.6 for the natural schedule and 8.3 for random promotion. Promoting the highest-surprisal token reaches 10.1 TPF but lowers accuracy to 73.3\%. With KL regularization, our criterion achieves the highest TPF and accuracy, at 10.1 and 78.9\%, respectively. It improves both measures over highest-surprisal promotion (8.9 TPF and 77.8\% accuracy) and raises TPF substantially over random promotion (7.5) while maintaining similar accuracy. These results support selecting tokens for how much they simplify the remaining predictions, rather than for their own surprisal alone.

\begin{table}[h]
\centering
\small
\caption{Effect of the token promotion strategy on LLaDA GSM8K, with and
without KL regularization. Accuracy is reported as a percentage.}
\label{tab:promotion-ablation}
\setlength{\tabcolsep}{7pt}
\begin{tabular}{lcccc}
\toprule
& \multicolumn{2}{c}{\textbf{Without KL}}
& \multicolumn{2}{c}{\textbf{With KL}} \\
\cmidrule(lr){2-3}\cmidrule(lr){4-5}
\textbf{Promotion strategy}
& TPF$\uparrow$ & Acc.$\uparrow$
& TPF$\uparrow$ & Acc.$\uparrow$ \\
\midrule
Natural schedule             & 7.6  & 74.9 & 6.3  & 78.0 \\
Random plausible token      & 8.3  & \textbf{75.1} & 7.5 & 78.7 \\
Highest-surprisal token     & 10.1 & 73.3 & 8.9  & 77.8 \\
Surprisal reduction (Eq.~\ref{eq:promotion-gain}) & \textbf{11.4} & 74.2
                            & \textbf{10.1} & \textbf{78.9} \\
\bottomrule
\end{tabular}
\end{table}

\subsubsection{KL Weight}
\label{app:beta}

\noindent\textbf{Sensitivity to the KL Weight.} We vary the KL regularization weight $\beta$ on LLaDA and report the results in Table~\ref{tab:app-beta}.  Without KL regularization, the model reaches the highest TPF but has the lowest accuracy on both GSM8K and MATH-500. Increasing $\beta$ to 1 raises \begin{wraptable}{r}{0.48\textwidth}
  \centering
  \small
  \vspace{-0.12in}
  \caption{Effect of the KL weight $\beta$ on LLaDA accuracy and TPF for GSM8K and MATH-500. \textbf{Bold} indicates the selected setting.}
  \label{tab:app-beta}
  \setlength{\tabcolsep}{5pt}
  \begin{tabular}{@{}ccccc@{}}
    \toprule
    & \multicolumn{2}{c}{\textbf{GSM8K}}
    & \multicolumn{2}{c}{\textbf{MATH-500}} \\
    \cmidrule(lr){2-3}\cmidrule(lr){4-5}
    $\boldsymbol{\beta}$
      & \textbf{Acc (\%)}$\uparrow$ & \textbf{TPF}$\uparrow$
      & \textbf{Acc (\%)}$\uparrow$ & \textbf{TPF}$\uparrow$ \\
    \midrule
    0          & 74.19 & 11.4 & 32.81 & 6.03 \\
    0.5        & 75.74 & 10.8 & 33.22 & 5.36 \\
    \textbf{1} & \textbf{78.94} & \textbf{10.12}
               & \textbf{34.40} & \textbf{4.96} \\
    2          & 79.10 & 8.1 & 34.40 & 4.61 \\
    \bottomrule
  \end{tabular}
  \vspace{-0.35in}
\end{wraptable}

accuracy from 74.19\% to 78.94\% on GSM8K and from 32.81\% to 34.40\% on MATH-500, while retaining 10.12 and 4.96 TPF, respectively. Raising $\beta$ further to 2 adds only 0.16 accuracy on GSM8K while reducing TPF from 10.12 to 8.1. On MATH-500, accuracy remains unchanged and TPF also falls.  We therefore use $\beta=1$ for its stronger balance between accuracy and parallel decoding.

\subsection{Qualitative Example}
\label{app:example}

In the following, we present the example of generation from our model and different baselines for LLaDA.

\begin{figure}[p]
\centering
\begin{tcolorbox}[enhanced, colback=white, colframe=black!55, boxrule=0.4pt, arc=0pt, left=4pt, right=4pt, top=3pt, bottom=3pt]
\small\textbf{Prompt:} \textit{A raspberry bush has 6 clusters of 20 fruit each and 67 individual fruit scattered across the bush. How many raspberries are there total?} \hfill Reference answer: 187
\end{tcolorbox}
\vspace{-2pt}
\begin{tcbraster}[raster columns=2, raster equal height, raster column skip=5pt, raster row skip=5pt]
\begin{exbox}{Base model (LLaDA-8B-Instruct)}{exbase}\footnotesize
To determine the total number of raspberries on the bush, we need to add the number of raspberries in the clusters to the number of individual raspberries scattered across the bush. Let's break this down step by step: 1.~Calculate the number of raspberries in the clusters: -- There are 6 clusters. -- Each cluster contains 20 raspberries. -- Therefore, the total number of raspberries in the clusters is: $6\times20=120$. 2.~Add the number of individual raspberries scattered across the bush: -- There are 67 individual raspberries scattered across the bush. -- Therefore, the total number of raspberries is: $120+67=187$. So, the total number of raspberries on the bush is $\boxed{187}$.
\exfoot{199}{199}{\exgood{Correct}}
\end{exbox}
\begin{exbox}{dParallel}{exdp}\footnotesize
To determine the total number of raspberries on the bush, we need to add the number of raspberries in the clusters and the number of individual raspberries scattered across the bush. 1.~First, calculate the number of raspberries in the clusters: -- There are 6 clusters, and each cluster contains 20 raspberries. -- Therefore, the number of raspberries in the clusters is: $6\times20=120$. 2.~Next, add the number of individual raspberries scattered across the bush: -- There are 67 individual raspberries scattered across the bush. -- Therefore, the total number of raspberries is: $120+67=187$. Thus, the total number of raspberries on the bush is $\boxed{187}$.
\exfoot{25}{188}{\exgood{Correct}}
\end{exbox}
\begin{exbox}{d3LLM}{exdt}\footnotesize
To determine the total number of raspberries on the bush, we need to calculate the number of raspberries in the clusters and the number of individual raspberries separately, and then sum these amounts. First, let's calculate the number of raspberries in the clusters: -- There are 6 clusters. -- Each cluster contains 20 raspberries. -- Therefore, the total number of raspberries in the clusters is: $6\text{ clusters}\times20\text{ raspberries per cluster}=120\text{ raspberries}$. Next, we add the number of individual raspberries: -- There are 67 individual raspberries scattered across the bush. -- Therefore, the total number of raspberries is: $120\text{ raspberries}+67\text{ raspberries}=187\text{ raspberries}$. Thus, the total number of raspberries on the bush is $\boxed{187}$.
\exfoot{23}{216}{\exgood{Correct}}
\end{exbox}
\begin{exbox}{\method{} (ours)}{exours}\footnotesize
To determine the total number of raspberries on the bush, we need to follow these steps: 1.~Calculate the number of raspberries in the clusters. 2.~Add the number of individual raspberries to the number of raspberries in the clusters. First, let's calculate the number of raspberries in the clusters: -- There are 6 clusters. -- Each cluster contains 20 raspberries. -- Therefore, the total number of raspberries in the clusters is: $6\times20=120$. Next, we add the number of individual raspberries to the number of raspberries in the clusters: -- There are 67 individual raspberries. -- Therefore, the total number of raspberries is: $120+67=187$. Thus, the total number of raspberries on the bush is $\boxed{187}$.
\exfoot{17}{202}{\exgood{Correct}}
\end{exbox}
\end{tcbraster}
\label{fig:app-example-59}
\end{figure}

\begin{figure}[p]
\centering
\begin{tcolorbox}[enhanced, colback=white, colframe=black!55, boxrule=0.4pt, arc=0pt, left=4pt, right=4pt, top=3pt, bottom=3pt]
\small\textbf{Prompt:} \textit{John wins an award at work. The award has a 1 time monetary reward of \$5000 and a 5\% raise in salary. If he makes 2000 a week, every week of the year and he got the award on the first day of the year how much money did he make that year?} \hfill Reference answer: 114{,}200
\end{tcolorbox}
\vspace{-2pt}
\begin{tcbraster}[raster columns=2, raster equal height, raster column skip=5pt, raster row skip=5pt]
\begin{exbox}{Base model (LLaDA-8B-Instruct)}{exbase}\footnotesize
To determine how much money John made that year, we need to consider both his one-time monetary reward and his annual salary after the 5\% raise. 1.~One-Time Monetary Reward: John receives a one-time monetary reward of \$5000. 2.~Annual Salary After Raise: John's weekly salary is \$2000. A 5\% raise in his salary means he will earn: \exbad{$2000\times0.05=1000$ dollars per week}. Therefore, his annual salary is: \exbad{$2000\times52=104000$ dollars}. 3.~Total Earnings for the Year: Adding the one-time monetary reward to his annual salary, we get: $5000+104000=109000$ dollars. Thus, the total amount of money John made that year is $\boxed{109000}$.
\exfoot{256}{255}{\exbad{Wrong}}
\end{exbox}
\begin{exbox}{dParallel}{exdp}\footnotesize
To determine how much money John made in the year, we need to consider both the one-time monetary reward and the percentage increase in his salary. 1.~One-time Monetary Reward: John receives a one-time monetary reward of \$5000. 2.~Percentage Increase in Salary: John receives a 5\% raise in his salary. His original salary is \$2000 per week. 3.~Total Annual Salary: There are 52 weeks in a year. Therefore, his annual salary is: \exbad{$2000\times52=104000$}. 4.~Total Earnings for the Year: Adding the one-time monetary reward to his total annual salary: $5000+104000=109000$. Thus, the total amount of money John made that year is $\boxed{109000}$.
\exfoot{45}{225}{\exbad{Wrong}}
\end{exbox}
\begin{exbox}{d3LLM}{exdt}\footnotesize
To determine how much money John made that year, we need to break down his earnings into two parts: the one-time monetary reward and his increased salary. 1.~One-Time Monetary Reward: John receives a one-time monetary reward of \$5000. 2.~Increased Salary: John receives a 5\% raise in his salary. His original salary is \$2000 per week. -- Calculate the raise amount: \exbad{$2000\times0.05=1000$} -- Add the raise to his original salary: $2000+1000=3000$ -- Calculate the total salary for the year: $3000\times52=156000$ 3.~Total Earnings for the Year: -- Add the one-time monetary reward to the total salary for the year: \exbad{$5000+16000=161000$}. Therefore, the total amount of money John made that year is \texttt{\textbackslash boxed\{161000} \exbad{[cut off at 256 tokens]}
\exfoot{46}{256}{\exbad{Wrong}}
\end{exbox}
\begin{exbox}{\method{} (ours)}{exours}\footnotesize
To determine how much money John made that year, we need to break down the problem into two parts: the one-time monetary reward and the 5\% raise in his salary. 1.~One-Time Monetary Reward: John receives a one-time monetary reward of \$5000. 2.~5\% Raise in Salary: John's original salary is \$2000 per week. A 5\% raise in his salary is: \exgood{$0.05\times2000=100$}. Therefore, his new salary is: \exgood{$2000+100=2100$}. There are 52 weeks in a year, so his annual salary is: $2100\times52=109200$. Now, we add the one-time monetary reward to his annual salary to find the total amount he made: $5000+109200=\boxed{114200}$.
\exfoot{28}{255}{\exgood{Correct}}
\end{exbox}
\end{tcbraster}
\label{fig:app-example}
\end{figure}

\end{document}